\documentclass{article} \usepackage{spconf} \newcommand{\figsize}{3.25}% Asilomar 
\usepackage{graphicx,psfrag}
\usepackage[cmex10]{amsmath}
\usepackage{amsfonts,amssymb,amsthm,latexsym,hyperref}
\usepackage{algorithm,algorithmic}

\usepackage{cite}
\usepackage[usenames,dvipsnames]{color}
\usepackage[normalem]{ulem} % provides strikeout command \sout

\usepackage{etoolbox}
\preto\subequations{\ifhmode\unskip\fi}% for subequations vspace problem

 \newcommand{\putFrag}[5]{\begin{figure}[t!]
                            \centering
                            #4
                            \includegraphics[width=#3in,#5,clip]{figures/#1.eps}
                            \vspace{-3mm}
                            \caption{#2}
                            \label{fig:#1}
                          \end{figure} }
 \newcommand{\putTable}[3]{\begin{table}[t!]
                            \centering
                            \caption{#2}
                            \vspace{0mm}
                            \scalebox{0.9}{#3} % adjust table size here
                            \vspace{0mm}
                            \label{tab:#1}
                          \end{table} }
 \newcommand{\capFrag}[2]{}
 \newcommand{\capTable}[2]{}

 \newcommand{\defn}{\triangleq}

 \newcommand{\Real}{{\mathbb{R}}}
 \newcommand{\Complex}{{\mathbb{C}}}

 \renewcommand{\eqref}[1]{(\ref{eq:#1})}

 \newcommand{\figref}[1]{Fig.~\ref{fig:#1}}
 \newcommand{\tabref}[1]{Table~\ref{tab:#1}}
 \newcommand{\secref}[1]{Section~\ref{sec:#1}}
 \newcommand{\Secref}[1]{Section~\ref{sec:#1}}

 \newcounter{comment}[section]
 
 \newcounter{texthead}[section]

\begin{document}

\setlength{\arraycolsep}{0.5mm}

%%%
\title{Deep Neural Networks for Radar Waveform Classification}

% for IEEEtrans.cls
%\author{Michael Wharton, Anne M. Pavy, and Philip Schniter%
%        \thanks{M. Wharton and P. Schniter are with the Dept.\ of Electrical and Computer Eng., The Ohio State University, Columbus OH 43210 (email: wharton.124@buckeyemail.osu.edu; schniter.1@osu.edu).
%                A. M. Pavy is with the Sensors Directorate, Air Force Research Lab, WPAFB, OH, 45433 (email: anne.pavy@us.af.mil).}
%        \thanks{This work was funded in part by the National Science Foundation under Grant 1539961.}
%       }

% for spconf.sty
\name{Michael Wharton$^\dagger$, Anne M. Pavy$^\ddag$, and Philip Schniter$^*$%
      \thanks{$^\dagger$Supported in part by NSF Grant 1539961}}
\address{$^\dagger \mbox{}^\ddag$Sensors Directorate, Air Force Research Lab, WPAFB, OH, 45433, \{michael.wharton.3,\,\,anne.pavy\}@us.af.mil\\
	         $^*$Dept. ECE, The Ohio State Univ., Columbus, OH, 43210, schniter.1@osu.edu}

\maketitle 

\begin{abstract}
We consider the problem of classifying radar pulses given raw I/Q waveforms in the presence of noise and absence of synchronization. We also consider the problem of classifying multiple superimposed radar pulses. For both, we design deep neural networks (DNNs) that are robust to synchronization, pulse width, and SNR. Our designs yield more than 100x reduction in error-rate over the current state-of-the-art. 
\end{abstract}

%\begin{IEEEkeywords}
%automatic target recognition, cognitive radar, classification, deep neural networks
%\end{IEEEkeywords}

%===============================================================================
\section{Introduction}

We consider the problem of classifying radar waveforms given as raw complex-valued time sequences. 
Waveform classification plays an important role in cognitive radar \cite{Haykin:SPM:06, Lunden:JSTSP:07} and other applications. 
For classifier design, we assume the availability of a dataset containing many examples of radar waveforms (i.e., sequences of complex-valued time-domain samples) with corresponding class labels in $\{1,\dots,K\}$; we do not assume any physical model. 
In this work, we focus on deep neural network (DNN) classifiers, as they have shown to be state-of-the-art in many related tasks.

The radar application imposes several unique challenges on DNN classifier design. 
First, radar pulse widths can span a very wide range (e.g., several orders-of-magnitude), even within a given class. 
%Most DNNs, however, assume a fixed input dimension. 
Second, radar pulses cannot be assumed to be time-synchronized in the observation window. 
Third, practical radar systems must operate over a huge range of signal-to-noise ratios (SNRs), including SNRs far below 0 dB. 
Fourth, the pulses are complex-valued, whereas most DNNs are designed for real-valued signals. 
All of these challenges make the classification of radar pulses quite different from typical classification problems to which DNNs are applied (e.g., image classification).
%yet, to be practical, must be handled simultaneously. 

%Several challenges are faced when designing DNN classifiers of raw radar waveforms, commonly referred to as ``pulses.''
%First, the pulse duration can span a very wide range (e.g., from hundreds to thousands of samples) even within a given class.
%Most DNNs, however, assume a fixed input dimension.
%Second, the pulses are complex-valued, whereas most DNNs are configured to support real-valued signals.
%Third, when applying the classifier, one cannot assume that the pulse will be time-synchronized, especially in passive sensing applications.
%Fourth, practical radar systems must operate over a wide range of signal-to-noise ratios (SNRs), including SNRs far below 0 dB.
%To be practical, all of these challenges must be simultaneously addressed.

Early approaches to radar waveform classification first converted the raw waveforms to hand-crafted, low-dimensional features, to which classic machine-learning techniques could be applied.
For example, \cite{Rigling:RAD:10} designed auto-correlation function (ACF) features of dimension 20 that were robust to time and frequency shifts, and trained a Fisher's linear discriminant classifier.
These ACF features were later used in \cite{Pavy:RAD:15} with a support vector machine (SVM) classifier and in \cite{Chakravarthy:RAD:20} with a shallow neural-network classifier.

It was recently demonstrated in \cite{Chakravarthy:RAD:20} that significantly better classification accuracy could be obtained by training a DNN to operate on raw time-domain radar waveforms in place of ACF features.
This is not surprising, since DNNs are able to learn features that outperform hand-crafted ones. 

Still, several assumptions were made in \cite{Chakravarthy:RAD:20} that limited both the performance and practicality of their design. 
First, the DNN input dimension in \cite{Chakravarthy:RAD:20} was chosen to be greater than the longest pulse in the training dataset.
We will show that it is better to truncate or pad the pulses to an optimized input length.
Second, the DNN in \cite{Chakravarthy:RAD:20} ignored the quadrature (Q) input to avoid complex-valued operations. 
We show that a well-designed complex-valued DNN has advantages over a real-valued DNN in this application.
Third, the training and test waveforms in \cite{Chakravarthy:RAD:20} were assumed to be time-synchronized, which is impractical. 
We train our DNN to be robust to time-asynchronous inputs.
Fourth, the DNN architecture in \cite{Chakravarthy:RAD:20} can be significantly improved, as we show.
%We make an effort to optimize our DNN architecture and training procedure. 
%We make an effort to optimize our DNN architecture (e.g., input dimension, depth, width, kernel size) and training procedure.% (e.g., batch size, learning rate).
Additionally, the training procedure can be improved via the use of data augmentation (i.e., random noise and delay realizations).
Fifth, \cite{Chakravarthy:RAD:20} assumed a single radar pulse, whereas we propose a DNN-based approach to multi-pulse radar waveform classification, which---to our knowledge---is the first in the literature. 

%===============================================================================
\section{Approach}\label{sec:architecures}

\subsection{Network Architecture}\label{sec:resnet}

Like in \cite{Chakravarthy:RAD:20}, we focus on feed-forward convolutional DNNs.
(Although we also experimented with recursive networks, we did not find the results to be competitive.)
%Feed-forward DNNs typically cascade many layers that each perform a similar sequence of operations: first convolve each input channel with a kernel, then linearly combine the convolution results into several output channels, then normalize each channel, and finally apply non-linear activation (e.g., ) to each channel.
%The kernel sizes and numbers of channels are design parameters that may vary across layers.
%
%Over several years of the ILSVRC competition \cite{ILSVRC:15}, variations of the aforementioned structure were proposed that led to increasingly better performance on image-classification tasks.
%One of the most famous outcomes of ILSVRC was the so-called deep residual network (ResNet) architecture \cite{He:ResNet}.
%Relative to standard feedforward DNNs, the ResNet adds ``skip connections'' 
%%(see \figref{resnet-radar}) 
%that facilitate the successful training of much deeper networks (e.g., 100s of layers) that consequently improve classification accuracy.
We evaluated 1D residual networks (ResNets) \cite{He:ResNet}, ResNeXts \cite{Xie:ResNeXt}, and DenseNets \cite{Huang:CVPR:17} due to their excellent performance on generic classification tasks. 
As we will discuss later, we found that the ResNet worked best for our data.
%In particular, we focus on deep residual networks (ResNets) \cite{He:ResNet} due to their excellent performance on related tasks.
%Although the DNNs considered were originally proposed for classification of images, they can be easily adapted to one-dimensional signals by changing the two-dimensional convolutions to one-dimensional convolutions and appropriately modifying the kernel sizes and numbers-of-channels.

%\putFrag{resnet-radar}{18-layer ResNet. Each convolutional layer is followed by batch-norm and nonlinear activation.  Here, ``1x$W$ conv $C$'' is a 1-dimensional convolution of kernel width $W$ and $C$ output channels.  The fully connected layer is essentially a convolutional layer with kernel size $W=1$.}{1.6}{}{trim=5 2 6 4}

%\textr{
%For single-pulse classification, we train with the standard cross-entropy (CE) loss. 
%For multi-pulse classification, we use the same network architecture, but train using binary cross-entropy (BCE) loss on each of the $K$ outputs.
%}

\subsection{Complex Arithmetic}\label{sec:complex}

%\textr{
%Most DNNs support only real-valued arithmetic, which is sufficient when the data is real-valued.
%But our radar waveforms are complex-valued. 
%The question is then how to best handle these complex waveforms.
%}

%Many approaches have been proposed to handle complex-valued signals with real-valued DNNs.
%Whether or not they are successful or not depends largely on the application.
%One of the simplest approaches is to feed the magnitude of the complex-valued signal to the DNN.
%Although this was successfully used in \cite{Wharton:ASIL:19} to classify synthetic aperture radar (SAR) images, we found that it did not work well for classification of phase-modulated radar waveforms.
%Another simple approach is to feed only the real part of the complex-valued signal to the DNN. 
%Although was successfully used in \cite{Chakravarthy:RAD:20} to classify phase-modulated radar waveforms, we show in the sequel that it can be improved upon.
%Another approach is to stack the real and imaginary parts of the waveform into a real-valued 2-row ``image'' and feed it to a real-valued DNN with 2-dimensional convolutions \cite{OShea:EANN:16}.
%Yet another approach is to feed the real and imaginary components into two separate input channels of a real-valued DNN, similar to how RGB image data is typically fed into three separate input channels.

Many approaches have been proposed to handle complex-valued signals with real-valued DNNs, however success depends largely on the application. 
Feeding the magnitude of the complex-valued signal to the DNN successfully classified synthetic aperture radar images in \cite{Wharton:ASIL:19}, but we found this to work poorly with our radar data. 
Another approach is to feed only the real part of the complex-valued signal to the DNN. 
Although this was used in \cite{Chakravarthy:RAD:20}, we show in the sequel that it can be improved upon.
Another approach is to stack the real and imaginary parts of the waveform into a real-valued 2-row ``image'' and feed it to a real-valued DNN with 2-dimensional convolutions \cite{OShea:EANN:16}.
Yet another approach is to feed the real and imaginary parts into two input channels of a real-valued DNN.%, similar to RGB image data.

An alternative is to design a complex-valued DNN. %as in \cite{Trabelsi:17, Cole:20}.
Such networks have led to improved performance in, e.g., audio classification \cite{Trabelsi:17} and magnetic resonance image (MRI) reconstruction \cite{Cole:20} tasks. 
To understand what we mean by a ``complex-valued DNN,'' consider the multiplication of a learnable parameter $k = k_{r}+ik_i\in\Complex$ with a feature $x = x_r+ix_i\in\Complex$, where $k_r,x_r\in\Real$ represent real parts, $k_i,x_i\in\Real$ represent imaginary parts, and $i\defn\sqrt{-1}$.
Such multiplications arise in the convolutional layers of DNNs.
The complex-valued multiplication of $k$ and $x$ can be written using four real-valued multiplications as
\begin{align} 
k x =(k_r x_r - k_i x_i ) + i(k_i x_r + k_r x_i )
\label{eq:complex_mult} .
\end{align}
The key point is that \eqref{complex_mult} is a two-input/two-output operation with only two learnable parameters: $k_r,k_i\in\Real$.
By contrast, a 2-channel real-valued DNN would implement
\begin{align} 
y_1 &= k_{11} x_1 + k_{12} x_2 \\
y_2 &= k_{21} x_1 + k_{22} x_2 
\label{eq:real_mult} ,
\end{align}
with $x_1\defn x_r$ and $x_2\defn x_i$, which involves four learnable parameters: $k_{11},k_{12},k_{21},k_{22}\in\Real$. 
By reducing the number of learnable parameters, we can reduce overfitting.

With regards to complex-valued activation functions, % (which are essentially two-input/two-output memoryless nonlinearities), there are many options, as discussed in \cite{Trabelsi:17, Cole:20}.
both \cite{Trabelsi:17, Cole:20} suggest that separately applying the ReLU function to the real and imaginary parts 
%\begin{align}
%\crelu(x) 
%&\defn \max\{0,x_r\}+i\max\{0,x_i\} \label{eq:crelu}\\
%&\quad\text{~for~} x=x_r+ix_i \nonumber
%\end{align}
outperforms other complex-valued activations in many applications, so we use this approach in our networks. %\eqref{crelu} in our network.

%For simplicity, we apply standard batch-norm separately to the real and imaginary outputs. 
Complex-valued implementations of batch-norm have also been developed (see, e.g., \cite{Cole:20}).
%%In our implementation, for simplicity, we apply standard batch-norm separately to the real and imaginary outputs.
For simplicity, we apply standard batch-norm separately to real and imaginary parts.% outputs.

\subsection{Noise Padding / Truncation / Delay}\label{sec:padding}

As mentioned earlier, our raw radar waveforms differ in duration from hundreds to thousands of samples.
But DNNs force us to choose a fixed input dimension for minibatch training.
The usual approach would be to set the input dimension equal to the longest sample and zero-pad the others as needed.

With noisy radar waveforms, however, it is more appropriate to noise-pad than to zero-pad, since in practice the test waveforms will be noise padded.
Thus, in \cite{Chakravarthy:RAD:20}, the waveforms were symmetrically padded with white Gaussian noise whose variance was chosen to match the noise variance in the original sample.
An example is shown in \figref{sync-padded}.

\putFrag{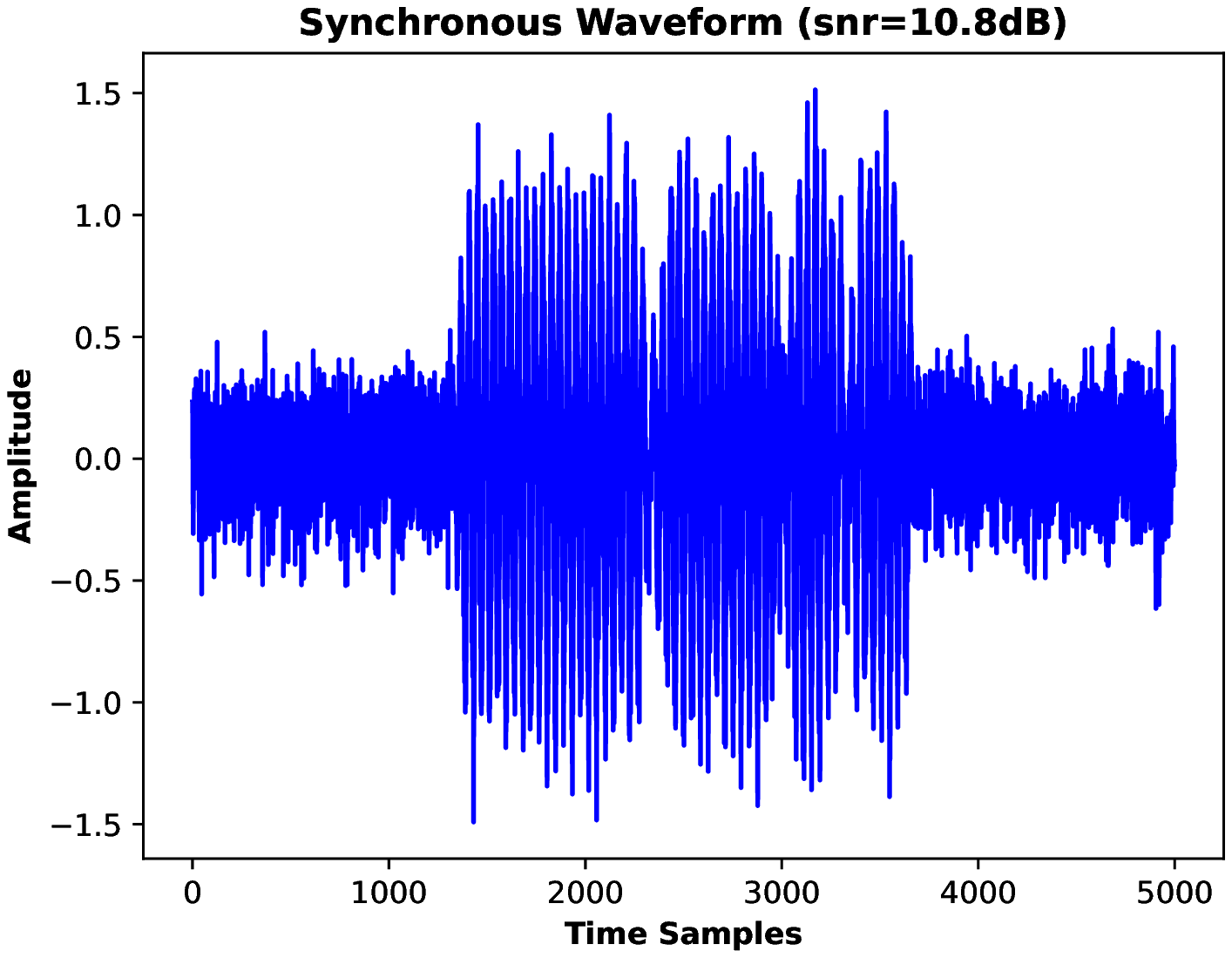}{Example synchronous noise-padded waveform at SNR=10.8dB, $N=2261$, and $D=5000$.}{\figsize}{}{trim=10 10 10 10}

There are several issues with the approach from \cite{Chakravarthy:RAD:20}.
First, as a consequence of symmetric padding, the padded waveforms will all be time-centered, i.e., synchronized.
But since time synchronization is not expected in practice, it is not advantageous to train on time-synchronized waveforms.
Second, noise-padding reduces SNR.
In particular, if a sample is $P\times$ longer after noise-padding, then its SNR will change by the factor $1/P$.
Third, \cite{Chakravarthy:RAD:20} padded each training sample with a \emph{fixed} noise waveform. 
As a result, the DNN might overfit to this particular noise realization. 
This latter problem is exacerbated by the very low SNRs encountered in radar.

Our approach is to noise-pad \emph{or} truncate the training waveforms as needed to obtain a fixed input length of $D$, where $D$ is optimized.
%This approach recognizes that noise-padding decreases SNR, while truncation discards discriminative features, and so the optimal approach must balance between these extremes.
%To avoid injecting time-synchronization bias into the training procedure, we randomly delay the training waveform whenever noise-padding or truncating that waveform.
To simulate asynchronous pulses, we randomly delay the pulse before noise-padding or truncating. 
We can describe this precisely using $N$ to denote pulse width and $U$ to denote a random integer uniformly distributed from $0$ and $|N\!-\!D|$.
When $N<D$, we noise-pad the front of the pulse using $U$ samples and noise-pad the back of the pulse using $D-N-U$ samples.
(See \figref{unsync-padded}.)
When $N>D$, we keep $D$ consecutive samples of the pulse starting at index $U$.
To avoid (over)fitting the DNN to particular training noise waveforms or delays, we draw new realizations of these quantities in every training minibatch (e.g., in the DataLoader of PyTorch \cite{Stevens:Book:20}).
This approach can be considered is a form of ``data augmentation'' \cite{Goodfellow:Book:16} that effectively increases the number of the training samples.
%Finally, we optimized the input dimension $D$ over a grid of possibilities, as described in the sequel.

%\putFrag{unsync-padded}{Example asynchronous noise-padded waveform at SNR=10.8dB, waveform length $N=2261$, DNN input length $D=5000$, delay $U=200$.}{\figsize}{}{trim=10 10 10 10}
\putFrag{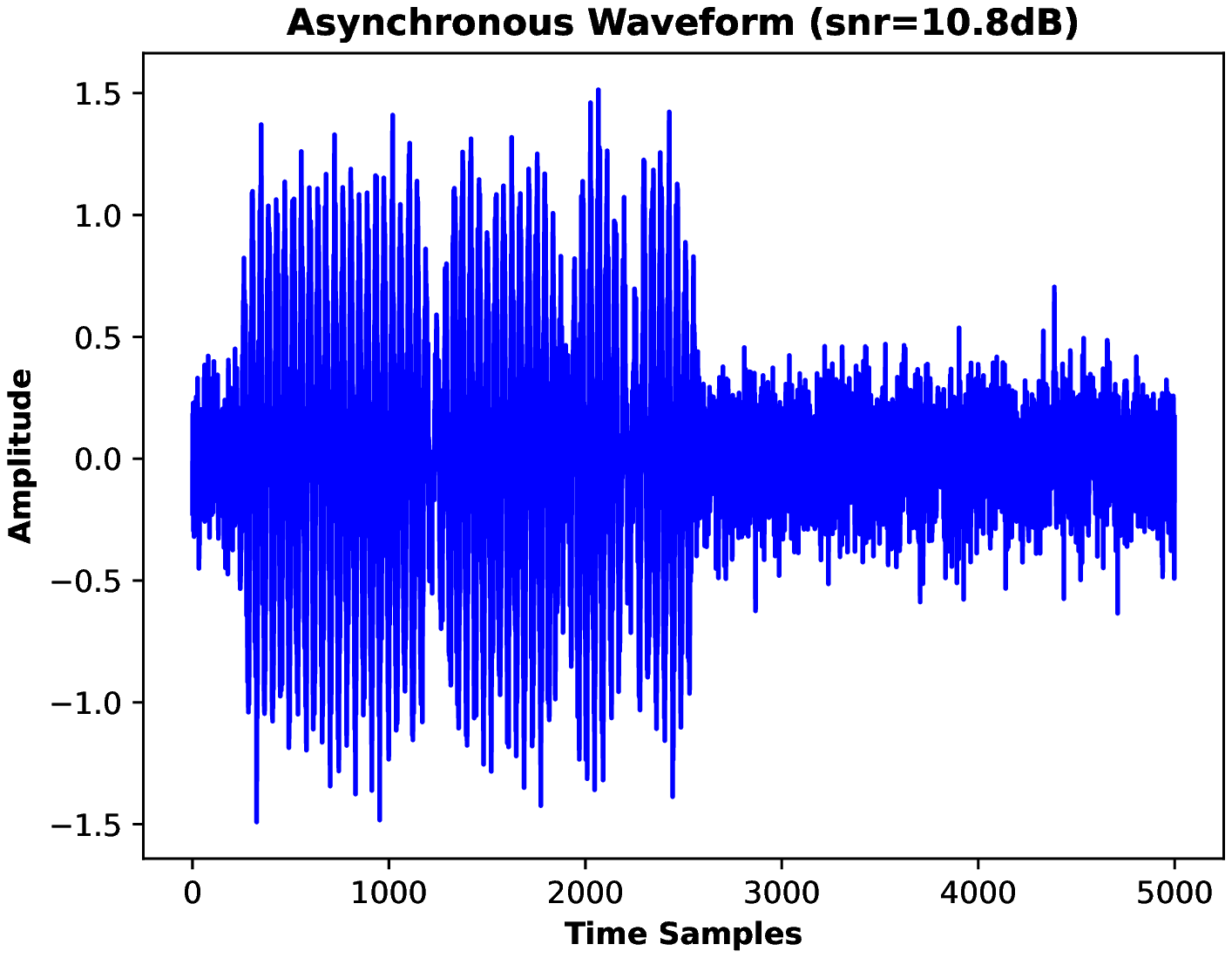}{Example asynchronous noise-padded waveform at SNR=10.8dB, $N=2261$, $D=5000$, $U=200$.}{\figsize}{}{trim=10 10 10 10}

%===============================================================================
\section{Experimental Results}

For our experiments, we use the SIDLE dataset, which was also used in \cite{Rigling:RAD:10,Pavy:RAD:15,Chakravarthy:RAD:20}.
This dataset contains 23 classes of phase-modulated radar pulses with 10000 examples of single pulses from each class.
(Please see \cite{Rigling:RAD:10,Pavy:RAD:15,Chakravarthy:RAD:20} for more details on the dataset.)
For a fair comparison to \cite{Chakravarthy:RAD:20}, we omitted classes 6 and 19--23 in the original dataset and used only the remaining $K\!=\!$~17 classes to train and test our DNN.
For these 17 classes, the dataset contains complex-valued waveforms with SNRs uniformly distributed between --12 and +12 dB. 
The modulation types and range of pulse widths are detailed in \cite{Chakravarthy:RAD:20}. 
For each experiment, we used a random subset of 80\% of the dataset for training and the remaining 20\% for testing.
%We used the Adam optimizer with default parameters $\beta_1=0.9$ and $\beta_2=0.999$.
In what follows, we discuss classification of single pulses in Sections~\ref{sec:baseline} and \ref{sec:improvements} and multiple pulses in \Secref{multi}.

\subsection{Baseline} \label{sec:baseline}

As a baseline, we first investigate the performance of the 9-layer real-valued DNN from \cite{Chakravarthy:RAD:20} on the task of single-pulse classification. 
As described earlier, this network uses an input dimension of 11000 and discards the imaginary part of the input.
To train and test the network, we used the synchronous noise-padding approach from \secref{padding} and illustrated in \figref{sync-padded}, and saw that the network converged to 0\% training error and 3.57\% test error, similar to what was reported in \cite{Chakravarthy:RAD:20}.

Next, to evaluate how well this DNN performs in the practical asynchronous setting, we re-trained it using the asynchronous noise-padding approach proposed in \secref{padding} and illustrated in \figref{unsync-padded}.
When testing with asynchronous data, we observed a test error of 18.29\%.
This relatively poor performance motivates the design of an improved DNN for asynchronous single-pulse classification.

\subsection{Improvements} \label{sec:improvements}

\noindent{\textbf{Architecture}: As a first step, we evaluated the ResNet, ResNeXt, and DenseNet architectures, which have more convolutional layers and fewer dense layers than the baseline DNN in \cite{Chakravarthy:RAD:20}. 
We configured each DNN using PyTorch's default parameters, but used 1D convolution. 
We trained and tested each network with the asynchronous noise-padding approach from \secref{padding}, and put the real and imaginary parts into two input channels, which we will show is superior to using just the real part. 
%The results are given in \tabref{dnn-arch-search} and show the ResNet performs best on asynchronous radar pulses, thus we optimize this architecture in our remaining experiments. 
The ResNet, ResNeXt, and DenseNet achieved test errors of 1.6\%, 10.5\%, and 2.8\%, respectively.
Thus, further experimentation focused on the ResNet.

%\putTable{dnn-arch-search}
%{DNN Architecture Comparison}{
%	\begin{tabular}{|c|c|c|c|}
%		\hline
%		DNN         & ResNet & ResNeXt & DenseNet \\ \hline
%		Test Error & \bf 1.6\% & 10.5\% & 2.8\% \\ \hline
%	\end{tabular}
%}
\medskip\noindent\textbf{ResNet}:
For fair comparison to the baseline DNN, we trained a 30-layer ResNet, but we used an input dimension of 11000 and only the real part of the waveform.
%As a first step, we swap the DNN from \cite{Chakravarthy:RAD:20} with a 30-layer ResNet, but still use input dimension 11000 and only the real part of the radar waveform.
%We configured the ResNet using PyTorch's default parameters, but with a one-dimensional kernel rather than a two-dimensional kernel.  
%The architecture is similar to the 18-layer ResNet shown in \figref{resnet-radar}. 
Training and testing this ResNet-30 using the asynchronous noise-padding approach from \secref{padding} gave 2.14\% test error, which significantly improves upon the 9-layer DNN from \cite{Chakravarthy:RAD:20}.

\medskip\noindent\textbf{Optimized input-length}:
To better understand how we might be able to improve the ResNet performance, we plot the classification outcome versus pre-padded pulse width and SNR in \figref{resnet30-errors}. 
The plot shows that pulses with both short length and low SNR are most likely to be misclassified.
This is consistent with the discussion in \secref{padding}, which described how noise-padding by factor $P$ changes the SNR by factor $1/P$.% causes the SNR to change by the factor $1/P$. 
%Thus, when the network input dimension is very large, short pulses will suffer a large SNR reduction that, when combined with an already low SNR, will likely result in misclassification.

\putFrag{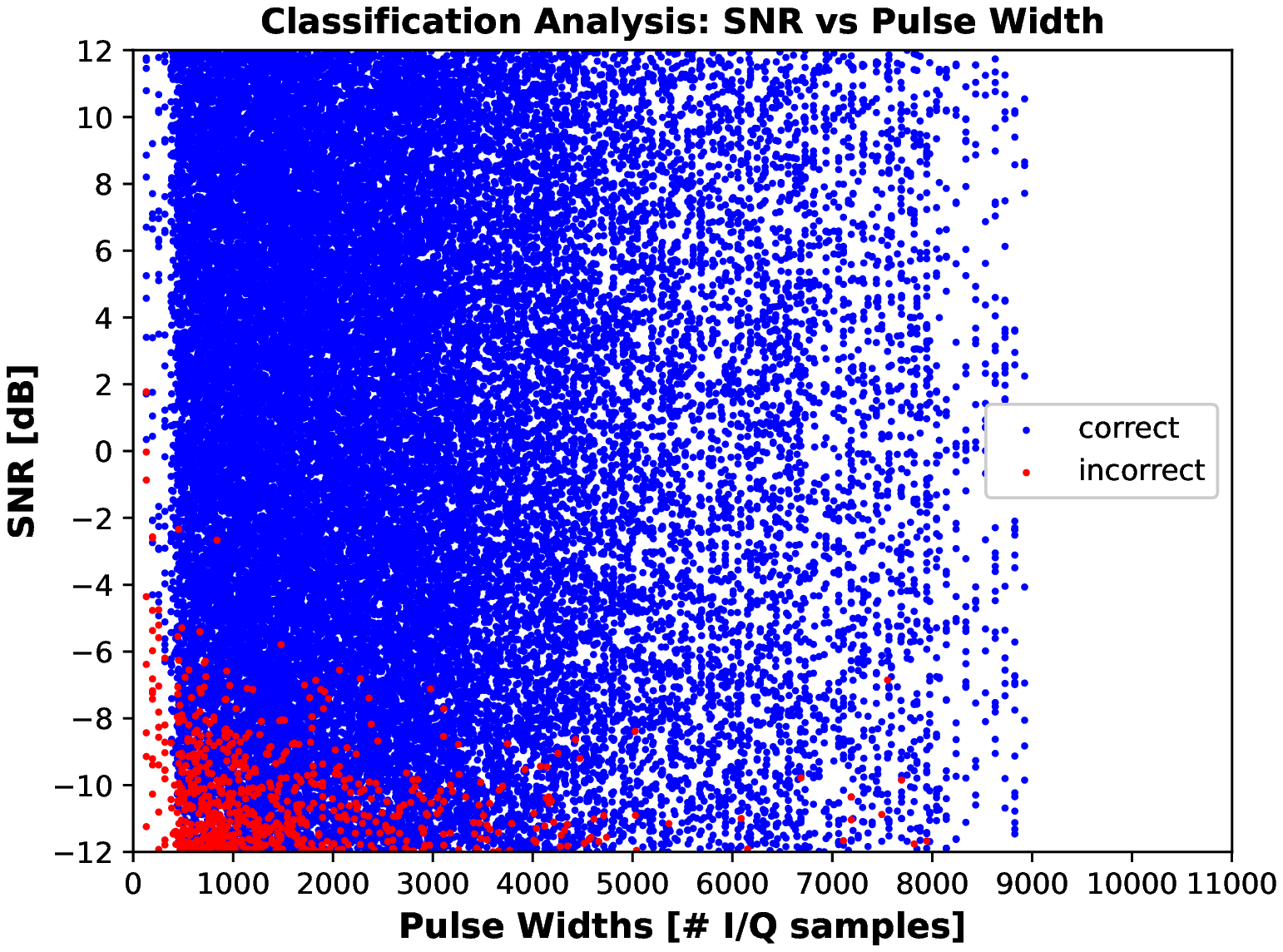}
        {Classification outcome (true=blue, false=red) of 11000-input ResNet-30 vs.\ pre-padded SNR and pulse length.}
        {\figsize}
        {}
        {trim=10 10 10 10}

To alleviate this problem, we considered reducing the network input dimension $D$ by either truncating or noise-padding each $N$-length raw waveform as needed.
%(as described in \secref{padding}).
\tabref{inputs-real-win-search} reports test error rate for several values of $D$.
The table shows that $D=3317$ was best among the tested values for the ResNet-30.
As a consequence of this $D$-optimization, the test error improved from 2.14\% to 1.32\%.

%\putTable{inputs-real-win-search}
%{ResNet-30 vs.\ Input Length}
%{
%	\begin{tabular}{|c||c|c|c|}
%		\hline
%		Input Length      & Test Error \\ \hline
%		1000    &  8.50\% \\ \hline
%		1821     & 2.16\%  \\ \hline
%		3317    & \bf 1.32\% \\ \hline
%		6040   & 1.35\% \\ \hline
%		11000  & 2.14\% \\ \hline
%	\end{tabular}
%}

%\putTable{inputs-real-11000}
%{Real-Valued Networks' Performance}
%{
%	\begin{tabular}{|c||c|c|c|}
%		\hline
%		Network                                      & Train $\&$ Test Data & Test Acc & Notes \\ \hline
%		\cite{Chakravarthy:RAD:20} & synchronous   &  96.43\% & overfit \\ \hline
%		\cite{Chakravarthy:RAD:20} & asynchronous &  81.71\%   & underfit \\ \hline
%		ResNet-18                                 & asynchronous &  93.80\% & underfit \\ \hline
%		ResNet-30                                & asynchronous & \bf 97.86\% & best \\ \hline
%		ResNet-50                                & asynchronous & 95.86\% & overfit\\ \hline
%	\end{tabular}
%}

\medskip\noindent\textbf{Complex-valued DNN}:
To improve the DNN further, we incorporate the complex-valued operations from \secref{complex} in the ResNet and refer to it as the ``$\Complex$ResNet.''
We also considered the 2-channel real-valued DNN from \eqref{real_mult}, which we call the ``IQ-ResNet.''
For a fair comparison, we adjusted the number of channels in the $\Complex$ResNet-30 and the IQ-ResNet-30 so that the number of trainable parameters is approximately equal to that in the (real-valued) ResNet-30. 
For all networks we report the number of real-valued trainable parameters (e.g., a complex-valued kernel of length $M$ has $2M$ trainable parameters).
The resulting test error rates are shown in \tabref{real-vs-complex} for  $D=3317$, which shows the improvement brought by the $\Complex$ResNet.

\putTable{inputs-real-win-search}
{ResNet-30 vs.\ Input Length}
{
	\begin{tabular}{|c||c|c|c|c|c|}
		\hline
		Input Length & 1000     & 1821      & 3317           & 6040    & 11000   \\ \hline
		Test Error      &  8.50\% &  2.16\% & \bf 1.32\%  &  1.35\% &  2.14\% \\ \hline
	\end{tabular}
}

\putTable{real-vs-complex}
{ResNet-30 vs.\ $\Complex$ResNet-30}{
	\begin{tabular}{|c||c|c|}
		\hline 
		Model               & Test Error & Trainable Parameters \\ \hline
		ResNet-30           & 1.52\%     & 1782193              \\ \hline
		IQ-ResNet-30        & 0.39\%     & 1782417              \\ \hline
		$\Complex$ResNet-30 & \bf 0.36\% & 1690189              \\ \hline
	\end{tabular}
}

\medskip\noindent\textbf{Network Fine-Tuning}:
As a final step, we performed an extensive fine-tuning of the $\Complex$ResNet.
This included a simultaneous search over
network depth,
network width,
kernel width,
batch size,
and
learning rate. 
For network and kernel width, we adjusted the 1st and 2nd layers respectively, and scaled the remaining layers 
%similarly to \figref{resnet-radar}.
in the same way as done in the default ResNets in PyTorch.
The test error (averaged over 100 random delays and noise realizations) and optimized network parameters of the fine-tuned $\Complex$ResNet are given in \tabref{complex-accs}.
A batch size of 512, learning rate of 0.001, and first-layer kernel width of 11 sufficed for all widths and depths.
Each network was trained for 90 epochs, with early stopping if the test loss did not improve for 15 consecutive epochs.
In the end, with fine-tuning, the test error rate dropped to 0.14\%.

\putTable{complex-accs}
{Fine-tuned $\Complex$ResNet Results vs.\ Network Depth}
{
	\begin{tabular}{|c||c|c|c|c|c|}
		\hline
		Model & Test Error & \# Parameters & Width & Kernel \\ \hline
%		$\Complex$ResNet-14   & 0.18\%  & 976049  & 16 & 11 \\ \hline
%		$\Complex$ResNet-18   & 0.17\%  & 1404209 & 16 & 9 \\ \hline
		$\Complex$ResNet-22   & 0.16\% &  7721041 & 32 & 11 \\ \hline
		$\Complex$ResNet-26   & 0.16\% &  1818161  & 16 & 7  \\ \hline
		$\Complex$ResNet-30   & \bf 0.14\% &  659233  & 8  & 9  \\ \hline
		$\Complex$ResNet-34   & 0.15\% &  670945  & 8  & 9  \\ \hline
		$\Complex$ResNet-38   & 0.16\% & 2228913 & 16 & 7 \\ \hline
	\end{tabular}
}

\subsection{Classification of multiple overlapping pulses} \label{sec:multi}

%We designed a now use the fine-tuned network architecture to classify multiple overlapping pulses. 
We now shift our focus to multi-pulse classification.
There, one observes a noisy superposition of $L$ shifted and scaled radar pulses and the goal is to determine whether a pulse is present or absent from each of the $K$ known classes.

To tackle this task, we simply retrained our fine-tuned $\Complex$ResNet to minimize binary cross-entropy (BCE) loss on each of the $K\!=\!$~17 outputs, rather than cross-entropy (CE) loss.
The resulting network outputs a real-valued logits vector, for which a positive entry in the $k$th location indicates that a pulse from the $k$th class is believed to be present, while a negative entry indicates that no pulse is believed to be present.

%Our experiments consider $L\in\{1,2,3,4\}$ overlapping waveforms. 
%A single network was trained by sampling $L$ uniformly, generating and adding $L$ asynchronous noiseless waveforms, and adding Gaussian noise to keep the SNRs of the multi-pulse waveforms between --12 and +12 dB.
%For a given $L$, multi-pulse waveforms were generated by first creating one asynchronous noise-padded waveform (as described in \secref{padding}) and then adding $L-1$ additional asynchronous noiseless zero-padded waveforms. 
%(using a noiseless version of the SIDLE dataset obtained from its authors).
%This way, the SNRs of the multi-pulse waveforms remained between --12 and +12 dB.
%A single network was trained by sampling $L$ uniformly over $\{1,2,3,4\}$.
%For a given $L$, we sum $L$ asynchronous noiseless waveforms, and add Gaussian noise to keep the SNRs of the multi-pulse waveforms between --12 and +12 dB. 

\newcommand{\ea}{E_{\text{abs}}}
\newcommand{\es}{E_{\text{sub}}}
There are two common ways to define the error-rate in multi-label classification: ``absolute error'' ($\ea$) refers to the error on the binary predictions, whereas ``subset error'' ($\es$) refers to the error on the $K$-ary prediction vector as a whole.
For i.i.d. binary errors, $\es=1-(1-\ea)^K\approx K\ea$ using the binomial approximation, which is accurate for small $\ea$.
%If the binary prediction errors are i.i.d., then $\es=1-(1-\ea)^K\approx K\ea$ using the binomial approximation, which is accurate for small $\ea$.

Our test error-rates for fixed $L\in\{1,2,3,4\}$ are presented in \figref{multilabel}.
There we see that both $\log \ea$ and $\log \es$ grow approximately linearly with $\log L$.
We also see that $\es \approx K\ea$.
Furthermore, we see that the single-pulse network (from \secref{improvements}) outperformed the multipulse network in the special case of $L\!=\!1$ test pulses, which is not surprising because it was trained for this special case.
Still, with 4 overlapping pulses, the multi-$L$ network achieves an absolute error of only 4.0\%.

\putFrag{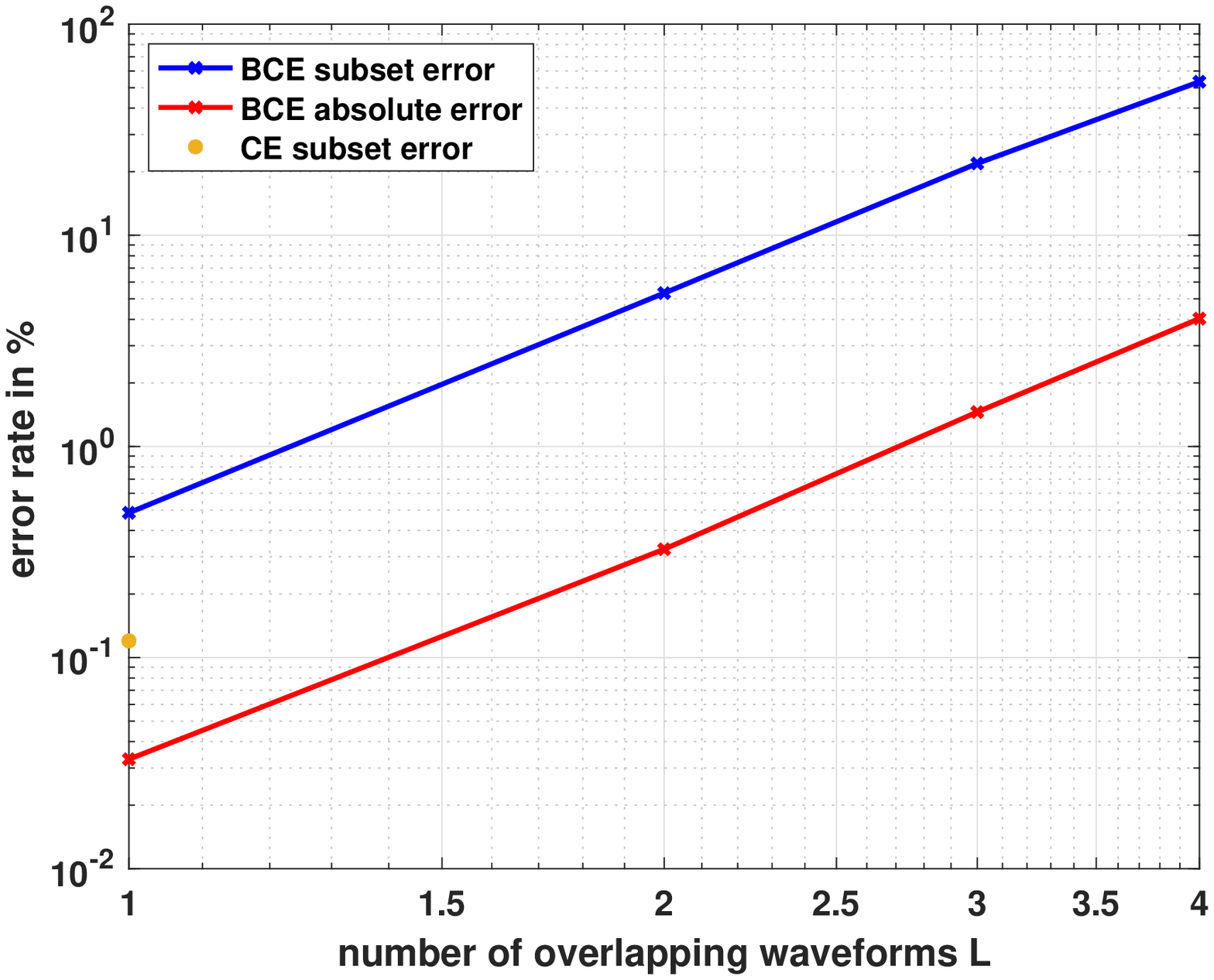}
        {Subset and absolute error-rate versus $L$ for the BCE-trained and CE-trained networks.}
        {3.0}
        {}
        {trim=0 0 0 0}

%===============================================================================
\section{Conclusion}

In this work, we considered the classification of complex-valued time-domain radar pulses specified by a large labeled dataset. 
For this purpose, we designed a complex-valued ResNet with optimized parameters, including width, depth, kernel width, batch size, and learning rate.
We also optimized the input dimension, necessitating either waveform truncation or noise-padding (as appropriate).
Our training procedure used random delays, random noise waveforms, and a wide range of SNRs for robustness and to avoid overfitting.
After fine-tuning, our network achieved a test error of 0.14\% on single-pulse classification of asynchronous SIDLE data, which is a 100x improvement over the previous state-of-the-art approach \cite{Chakravarthy:RAD:20}, which achieved 18.29\%.
When trained for multi-pulse classification, our network obtained an absolute error of 4.0\% with 4 overlapping pulses.
For future work, it would be interesting to investigate alternative approaches to multi-pulse classification, such as formulating it as a time-domain object detection problem. 

%``Noise class'' 
%- can we this
%- how does this affect low-snr samples
%``Object detection''

%%%%%%%%%%%%%%%%%%%%%%%%%%%%%%%%%%%%%%%%%%%%%%%%%%%%%%%%%%%%
\clearpage
\bibliographystyle{ieeetr}
\bibliography{macros_abbrev,books,machine,sparse,radar} % in /bib of group SVN

\end{document}